\documentclass[letterpaper, 10 pt, conference]{ieeeconf}  

\IEEEoverridecommandlockouts                              

\usepackage{graphicx}
\usepackage[export]{adjustbox}
\usepackage{float}
\usepackage{adjustbox}
\usepackage{comment}

\usepackage{xcolor}
\usepackage{amsmath,amssymb}
\usepackage{tabu, makecell, graphicx, booktabs}
\usepackage{threeparttable}
\usepackage[tight,footnotesize]{subfigure}

\title{\LARGE \bf
SANGO: Socially Aware Navigation through Grouped Obstacles}

\author{Rahath Malladi$^{*}$, Amol Harsh, Arshia Sangwan, Sunita Chauhan, and Sandeep Manjanna
\thanks{All the authors are at Plaksha University, Mohali, India
        {\tt\small correspondences: msandeep.sjce@gmail.com}}
}

\begin{document}

\maketitle
\thispagestyle{empty}
\pagestyle{empty}

\begin{abstract}

This paper introduces SANGO (Socially Aware Navigation through Grouped Obstacles), a novel method that ensures socially appropriate behaviour by dynamically grouping obstacles and adhering to social norms. Using deep reinforcement learning, SANGO trains agents to navigate complex environments leveraging the DBSCAN algorithm for obstacle clustering and Proximal Policy Optimization (PPO) for path planning. The proposed approach improves safety and social compliance by maintaining appropriate distances and reducing collision rates. Extensive experiments conducted in custom simulation environments demonstrate SANGO's superior performance in significantly reducing discomfort (by up to
$83.5\%$), reducing collision rates (by up to $29.4\%$), and achieving higher successful navigation in dynamic and crowded scenarios. These findings highlight the potential of SANGO for real-world applications, paving the way for advanced socially adept robotic navigation systems. 

\end{abstract}

\section{INTRODUCTION} 

\begin{figure}[h]
    \centering
    \includegraphics[width=0.95\linewidth]{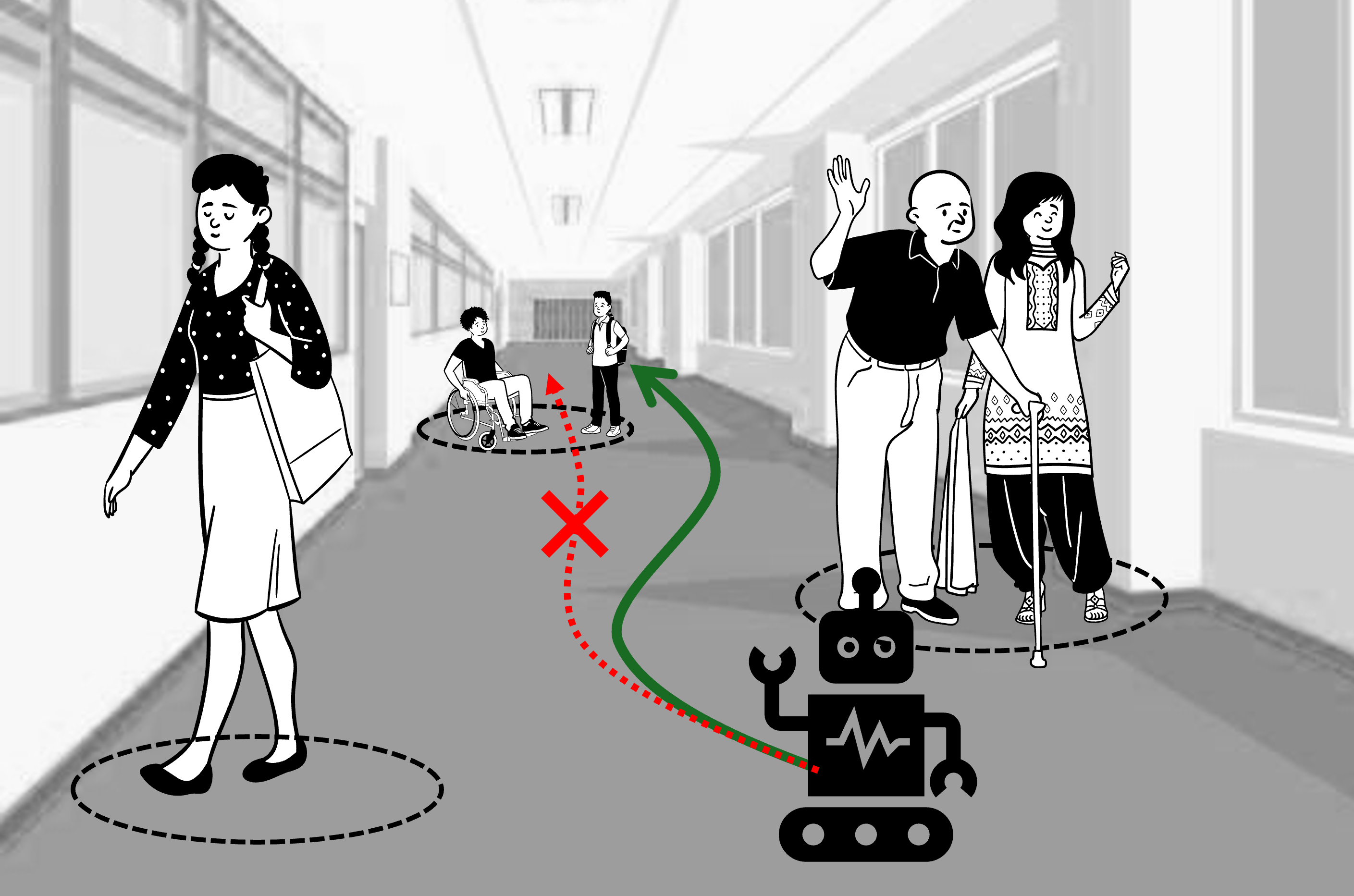}
    \caption{An illustration of the proposed social navigation technique (SANGO). The agent groups the dynamic obstacles (humans in this case) into groups and uses this grouping to plan a path to the goal location.}
    \label{fig:intro}
\end{figure}

In the rapidly advancing field of robotics, integrating robotic agents into everyday human environments requires significant progress in socially aware robotic navigation. This paper introduces SANGO (Socially Aware Navigation through Grouped Obstacles), a method for path planning in dynamic environments. SANGO generates socially appropriate behaviour by dynamically grouping obstacles and ensuring safety while adhering to social norms. We employ deep reinforcement learning techniques to train and test our agent in complex environments, validating our approach. As shown in Figure~\ref{fig:intro}, the robot groups human obstacles according to their observed interactions and plans its path accordingly.

The field of Socially Aware Robot Navigation(SARN) has garnered significant attention in recent years. As robots become more prevalent in daily life, their ability to navigate socially and efficiently is critical. Early research focused primarily on basic collision avoidance and path planning. However, integration of social norms and human comfort has become a focal point, necessitating a shift from traditional methods to more advanced learning-based approaches. One of the first articles on robot social navigation discusses the integration of social norms and behaviours into robot navigation systems, aiming to make robots navigate more naturally and comfortably in human-populated environments~\cite{luber2012socially}. Many recent studies have emphasized the importance of human-aware navigation, categorizing research into comfort, naturalness, sociability, and technology-driven approaches~\cite{KRUSE20131726, mavrogiannis2023core} and have expanded on these principles by exploring deep reinforcement learning (DRL) methods for navigation~\cite{Gao_and_Huang, 9409758, konar2021learning, baghi2022sesno}. These studies differentiate between local obstacle avoidance, indoor navigation, multi-robot systems, and social navigation, underlining the versatility of DRL in addressing various challenges.

The incorporation of human motion prediction and multi-behaviour navigation is crucial for enhancing social acceptability. The authors of \cite{Chik_Yeong_Su_Lim_Subramaniam_Chin_2016} discuss hierarchical navigation tasks, emphasizing the integration of high-level decision-making with reactive low-level obstacle avoidance. This approach ensures that robots can adapt to dynamic human environments, predicting movements and adjusting navigation behaviours accordingly. In \cite{wang2022groupbasedmotionpredictionnavigation}, the authors present a Model Predictive Control framework (G-MPC) that improves robot navigation by leveraging group-based motion prediction in crowded environments. Our research builds on these insights by leveraging Proximal Policy Optimization (PPO) \cite{schulman2017proximal} to train agents in environments that simulate real-world complexities. By dynamically grouping obstacles and navigating through them, our agent not only avoids collisions, but also maintains socially acceptable distances and behaviours. This dynamic grouping is critical in environments with high human traffic, where traditional static obstacle avoidance methods fall short. The major contributions of this paper include: 1) a novel deep reinforcement learning based socially aware navigation algorithm using a dynamic grouping of the obstacles (SANGO), and 2) custom-built simulation environments, Modular Socially Aware Navigation Gym (MOSANG) and Complex Open Gym (COG), to visualize and validate social navigation techniques.

\section{METHODOLOGY}

In this study, we introduce SANGO, a methodology meticulously designed to train an agent in dynamic and complex environments. SANGO employs the Density-Based Spatial Clustering of Applications with Noise (DBSCAN) algorithm as a grouping mechanism to form obstacle clusters and derive rewards based on the agent's navigation around these groups. Using the PPO algorithm, the agent is trained to achieve optimal performance by incorporating static and dynamic obstacles, generating stochastic obstacle trajectories, and utilizing sophisticated reward functions. This approach fosters socially aware navigation, resulting in a robust SANGO agent adept at navigating social environments.

\subsection{Grouping Reward, DBSCAN and PPO Algorithms}

The reward corresponding to dynamic obstacle grouping is derived using the DBSCAN algorithm, which clusters data points based on density, identifying regions of high density (clusters) separated by regions of low density (noise). The core idea is to find points closely packed together and mark them as clusters, while points lying alone in low-density regions are considered noise, which are then treated as individual dynamic obstacles. In contrast to algorithms such as K-means or hierarchical clustering which impose restrictions like predefined cluster numbers or lack of adaptability to dynamically shifting obstacles, DBSCAN's ability to discover clusters of arbitrary shape and size without requiring the number of clusters to be specified a priori makes it ideal for dynamically grouping moving obstacles. 

At each time step, DBSCAN is applied to obstacles within a range of 6-8 grid units (6-8 meters) in the COG environment and 60-80 grid units (also 6-8 meters) in MOSANG. With the environment represented in two dimensions and dynamic obstacles treated as point entities, clustering is based solely on their positions. The positions of these groups are continuously updated while within the agent's vicinity, and if a group moves out of range for a set duration, it is removed from memory. This dynamic grouping approach allows the agent to navigate around clusters of obstacles and to receive rewards for maintaining socially coherent navigation patterns. In our current approach, we assume that the agent has a specified local view of the world, akin to a LiDAR scan, which enables the agent to dynamically group the dynamic obstacles to adapt its navigation strategies. 

Unlike algorithms such as DDPG or TRPO, PPO strikes a balance between exploration and exploitation while maintaining stable policy updates, making it well-suited for our environments with frequent state transitions and stochasticity. Moreover, the integration of DBSCAN with PPO in our framework introduces an interesting synergy where dynamic grouping directly informs the agent’s path-planning decisions—allowing for contextually aware navigation that prioritizes both safety and social compliance. This approach provides a novel way to train the agent to navigate in socially coherent patterns, demonstrating the effectiveness of dynamic obstacle grouping in complex environments.

\subsection{Agent Architecture}

The agent is built using the standard PPO architecture, leveraging the MultiInputPolicy from the Stable Baselines3~\cite{stable-baselines3} library, with a learning rate of 0.0006, a discount factor of 0.97, and varying horizons based on the experiment (discussed further in the Experimentation section). PPO is chosen for its ability to effectively balance exploration and exploitation, making it suitable for complex environments with high-dimensional observation spaces. The MultiInputPolicy allows the agent to process various inputs, enhancing its capacity to learn from the various elements present in the environment.

\subsection{Reward Function}

The reward function (\ref{eqn:Reward_Function}) is designed to guide the agent toward achieving socially acceptable navigation behaviours. These rewards are formulated to incentivize the agent to navigate efficiently while maintaining social norms by avoiding collisions and respecting personal space. The specific mathematical formulation is as follows:

\begin{equation} \label{eqn:Reward_Function}
    \textbf{\emph{R}} = 
    \begin{cases}
        -30 & \text{if $d_t(\phi_t, {\psi_t}^{(i)}) = 0$} \\
        -20 & \text{if $d_t(\phi_t, {\varphi}^{(i)}) = 0$} \\
        -20 & \text{if $d_t(\phi_t, {\varpi}^{(i)}) = 0$} \\
        -50 & \text{if $d_t(\phi_t, {\psi_{{c}_t}}^{(i)}) = 0$} \\
        -15 & \text{if $d_t(\phi_t, {\varpi}^{(i)}) \leq \eta_1$} \\
        -\frac{20}{d_t(\phi_t, {\psi_t}^{(i)})} & \text{if $d_t(\phi_t, {\psi_t}^{(i)}) \leq \eta_2$} \\
        -e^{\frac{3}{d_t(\phi_t, {\psi_{{b}_t}}^{(i)})}} & \text{if $d_t(\phi_t, {\psi_{{b}_t}}^{(i)}) \leq \eta_3$} \\
        -\delta_t(\phi_t, \phi_{t-1}) \times \zeta & \text{$\forall \delta_t(\phi_t, \phi_{t-1})$} \\
        -2500 & \text{if $t \geq \tau$} \\
        3000 & \text{if $(\phi_t = \varepsilon)$} \\
        -1 & \text{$\forall t \in [0, \tau]$}
    \end{cases}
\end{equation}

\begin{itemize}
    \item \textbf{$\phi_t$}: Agent coordinates at time $t$.
    \item \textbf{$\psi_t$}: Dynamic obstacle coordinates at time $t$.
    \item \textbf{$\psi_{c_t}$}: Core dynamic obstacle coordinates post-grouping at time $t$.
    \item \textbf{$\psi_{b_t}$}: Boundary dynamic obstacle coordinates post-grouping at time $t$.
    \item \textbf{$\varphi$}: Static obstacle coordinates.
    \item \textbf{$\varpi$}: Environmental boundary coordinates.
    \item \textbf{$\varepsilon$}: Goal coordinates.
    \item \textbf{$\eta_1, \eta_2, \eta_3$}: Distance thresholds for penalty computation:
        \begin{itemize}
            \item \textbf{$\eta_1$}: Agent-to-boundary.
            \item \textbf{$\eta_2$}: Agent-to-dynamic obstacle.
            \item \textbf{$\eta_3$}: Agent-to-group boundary obstacle.
        \end{itemize}
    \item \textbf{$\zeta$}: Scaling factor for goal-directed progress ($\zeta = 4.688$ in this study).
    \item \textbf{$\tau$}: Task time horizon.
    \item \textbf{$d_t(A,B)$}: Euclidean distance (L2 norm) between $A$ and $B$ at time $t$.
    \item \textbf{$\delta_t(A_t, A_{t-1})$}: Difference in Euclidean distance between agent $A$ and the goal at time $t$ and $t-1$.
\end{itemize}

\begin{figure*}[!ht]
	\centering
	\subfigure[World 1 featuring 75 dynamic obstacles, with closely positioned obstacles grouped together and shown in cyan colour.]{ \includegraphics[height=0.21\textheight, keepaspectratio]{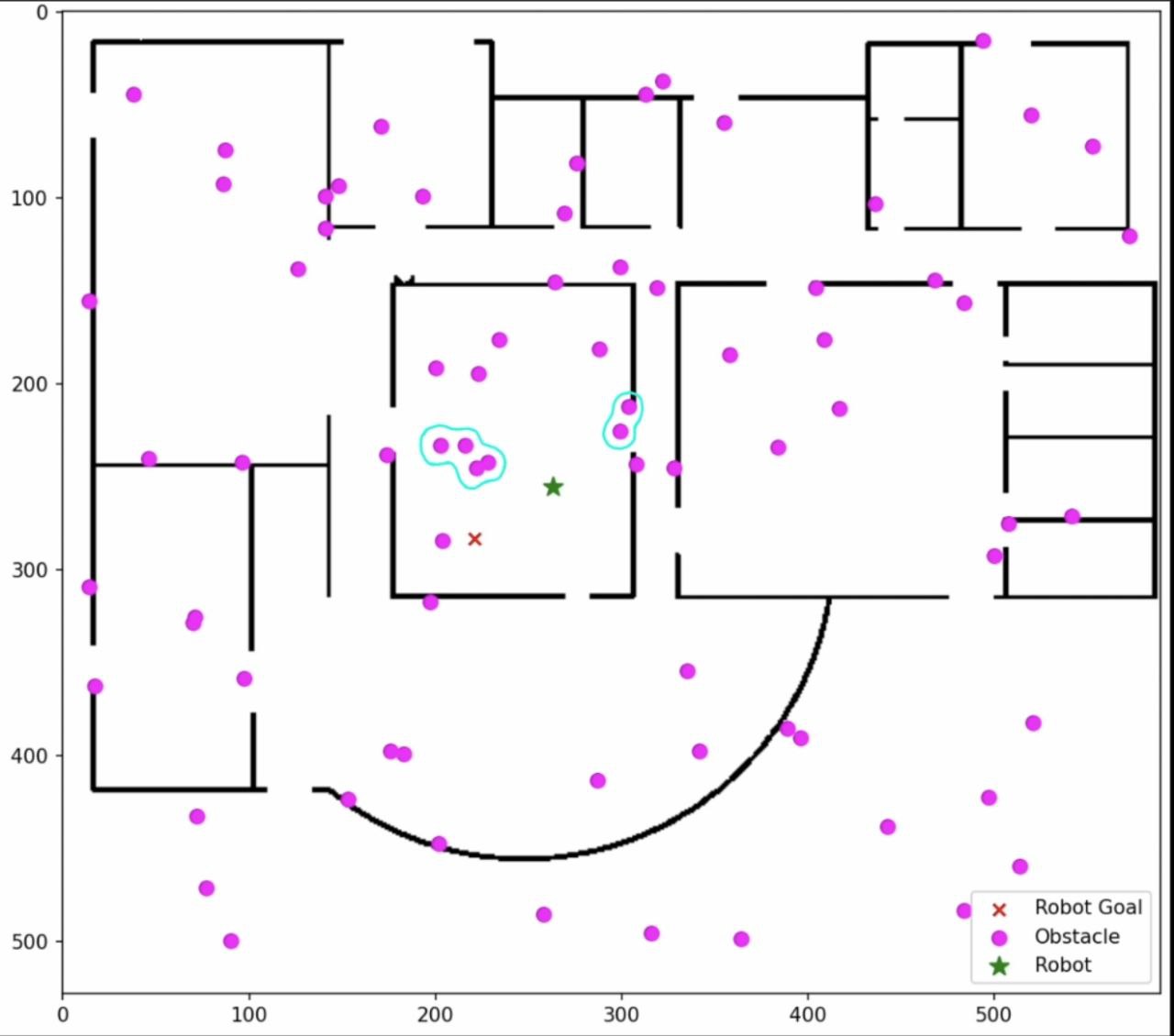}\label{fig:world1}}
	\subfigure[World 2 with 30 dynamic obstacles]{ \includegraphics[height=0.21\textheight, keepaspectratio]{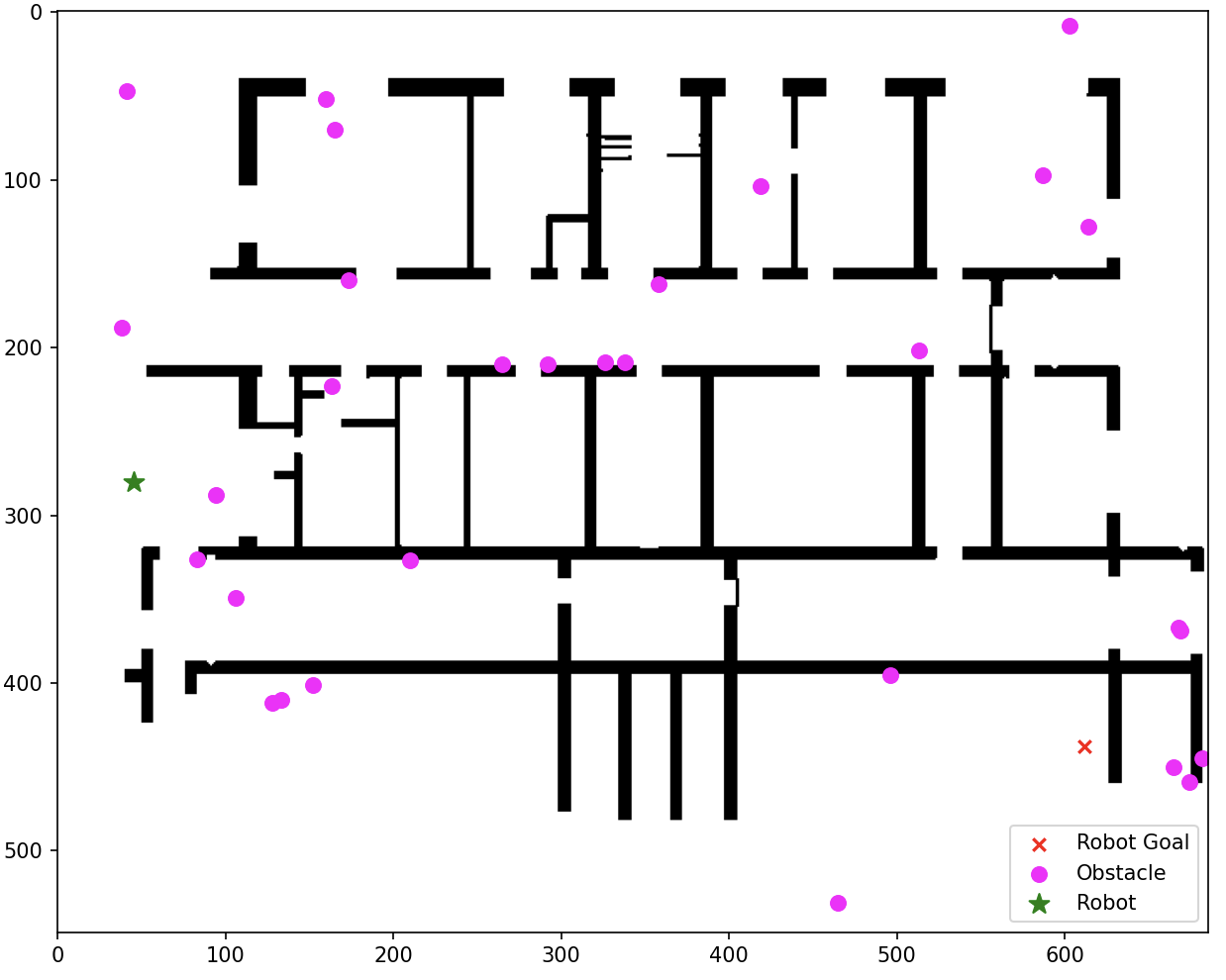}\label{fig:world2}}
	\subfigure[World 3 with 10 dynamic obstacles]{ \includegraphics[height=0.21\textheight, keepaspectratio]{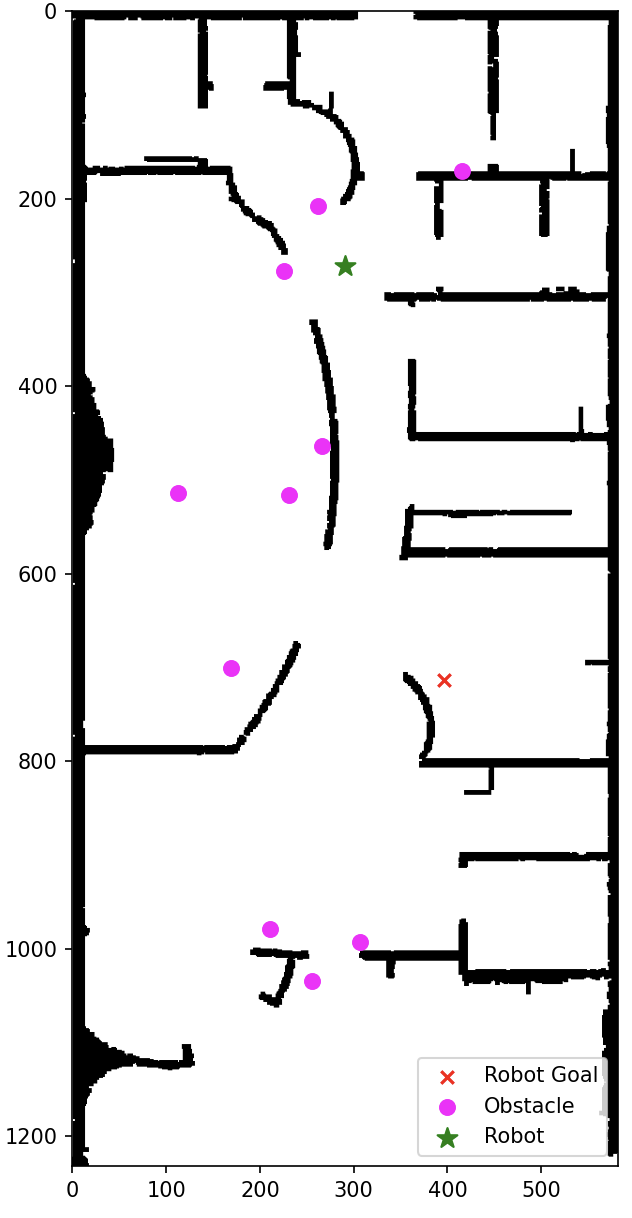}\label{fig:world3}}
    \caption{Example scenarios with dynamic obstacles that can be used to train agents in diverse scenarios.  Magenta circles represent the obstacles, the green star represents the robot, and the red cross represents the goal.}
	\label{fig:sim_worlds}
\end{figure*}

\textbf{The Design:} The reward function promotes safe, socially-aware navigation by penalizing collisions ($\psi_t$, $\varphi$, $\varpi$), proximity to obstacles, and intrusions into groups ($\psi_{c_t}$, $\psi_{b_t}$) while rewarding efficient goal-reaching behaviour. Time efficiency is encouraged through live and timeout penalties and goal achievement is strongly incentivized with a large reward. All calculations are performed in the global frame.

\section{ENVIRONMENTS \& EXPERIMENTS}

To train the agent, we developed two custom simulation environments: MOSANG and COG, where MOSANG focuses on training the agent to navigate complex worlds filled with dynamic obstacles, while COG trains the agent to manoeuvre through chaotic indoor spaces with a mix of dynamic and static obstacles.

\subsection{MOSANG - Modular Socially Aware Navigation Gym}

MOSANG is a versatile gym environment that is capable of importing blueprints and using them as training maps. This environment initializes a 2D grid world from blueprint images, converting spatial layouts into matrix grids that represent free spaces, boundaries, and obstacles. 
 
The simulator exposes tunable parameters, such as the pixel threshold, the dilation factor, and the area filters, to customize the blueprint adaptation. Figure~\ref{fig:sim_worlds} presents three example scenarios with dynamic obstacles that can be used to train agents in diverse scenarios. The entire code base is in Python, with \texttt{matplotlib} for live animation, providing a lightweight platform for effective experimentation.

Key simulation parameters include the number of dynamic obstacles, episode length, total episodes, and total training steps. The action space for the agent includes the king's moves, with integers from 0 to 8 corresponding to the 9 actions: up, down, left, right, right-up, left-up, right-down, left-down, and staying still. Dynamic obstacles, which move using the A* pathfinding algorithm with noise, have their paths recorded to ensure collision-free trajectories. Visualization uses \texttt{matplotlib} with magenta circles for obstacles, a green star for the robot, and a red cross for the goal, aiding in training evaluation.

\subsection{COG - Complex Open Gym}

\begin{figure}[!h]
    \centering
    \includegraphics[width=0.7\linewidth]{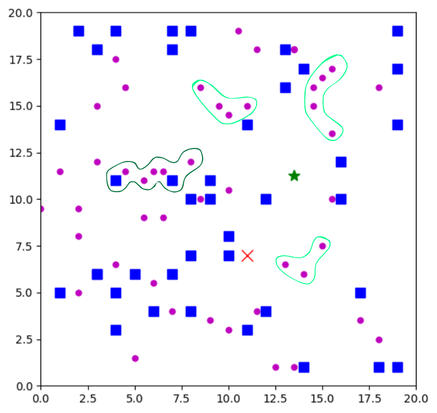}
    \caption{An Experiment Scenario: SANGO Agent (green star) navigating a COG environment with 40 static and 50 dynamic obstacles towards the goal (red cross). Cyan groups represent obstacles currently in the agent's vicinity, while the dark green group was nearby in the previous time step.}
    \label{fig:environment}
\end{figure}

These simulations are designed to emulate chaotic and complex indoor spaces, featuring multiple static and dynamic obstacles. Constructed using the Gymnasium library of OpenAI~\cite{brockman2016openaigym}, it presents a robust training and testbed for our agent's social navigation requirements and capabilities. As illustrated in Figure~\ref{fig:environment}, the agent is represented by a green star, the goal position by a red cross, static obstacles by blue squares, and dynamic obstacles by pink circles in motion.

The agent's state is encoded as a dictionary comprising its position, the positions of static obstacles, and the distances and angles of dynamic obstacles relative to the agent. These state values are encapsulated using the `Box` type from the Gym library, and adjusted according to the grid size. The agent uses the king's moves to navigate. The Euclidean distance between the agent and dynamic obstacles is computed for proximity assessment and angles are derived using the arctangent function based on their relative positions. The reward function is consistent with the methodology section, utilizing PPO to train the agent for socially aware navigation. Both our environments support training RL agents with algorithms from the Stable Baselines3 library as well as any custom implementation.

\subsection{Learning Scenarios}

The experiments were conducted using World 1 (Figure~\ref{fig:world1}), with various instantiations of the following configurations in both MOSANG and COG simulations:
\begin{itemize}
    \item \textbf{Simple}: Agent navigating with 3 dynamic obstacles in MOSANG and agent navigating with 10 dispersed static obstacles and 10 dynamic obstacles in COG.
    \item \textbf{Medium}: Agent navigating with 30 dynamic obstacles in MOSANG and agent navigating in a complex environment with 40 dispersed static obstacles and 30 dynamic obstacles in COG.
    \item \textbf{Complex}: Agent navigating with dynamic grouping and 30 dynamic obstacles in MOSANG and agent navigating in an environment featuring 40 dispersed static obstacles and 50 dynamic obstacles in COG (Figure~\ref{fig:environment}).
\end{itemize}

\subsection{Environment and Obstacle Initialization}

At the beginning of each episode, both static and dynamic obstacles are instantiated within the environment. Static obstacles remain fixed, while dynamic obstacles follow predefined courses, ensuring continuous movement throughout the episode. As dynamic obstacles reach their goals, new courses are spawned to maintain their motion until the episode's horizon. This dynamic setup ensures that the agent encounters a realistic and challenging navigation scenario.

\subsection{Trajectory Generation}

The trajectories of dynamic obstacles are generated randomly using a combination of the modified Social Force Model (SFM) \cite{SFM}, Optimal Reciprocal Collision Avoidance (ORCA) \cite{ORCA}, and Noisy A* algorithms. Each algorithm is utilized as required within the episode to simulate realistic pedestrian movements and obstacle interactions:

\begin{itemize}
    \item \textbf{Social Force Model}: Simulates the social forces acting on pedestrians, influencing their movements based on personal space and goal-directed behaviour.
    \item \textbf{Optimal Reciprocal Collision Avoidance}: Ensures collision-free trajectories by predicting and avoiding potential collisions with other moving obstacles.
    \item \textbf{Noisy A*}: Adds stochasticity to the A* pathfinding algorithm, introducing variability in obstacle trajectories to simulate real-world unpredictability.
\end{itemize}

\subsection{Metrics}

In the spectrum of experiments conducted to train the agent in various environments, with differing dynamic obstacle densities and complexities, the following metrics were employed to evaluate the agent's performance while learning:

\begin{figure}[!h]
    \centering
    \includegraphics[trim={0, 0.4cm, 0, 0.525cm}, clip, width=\linewidth]{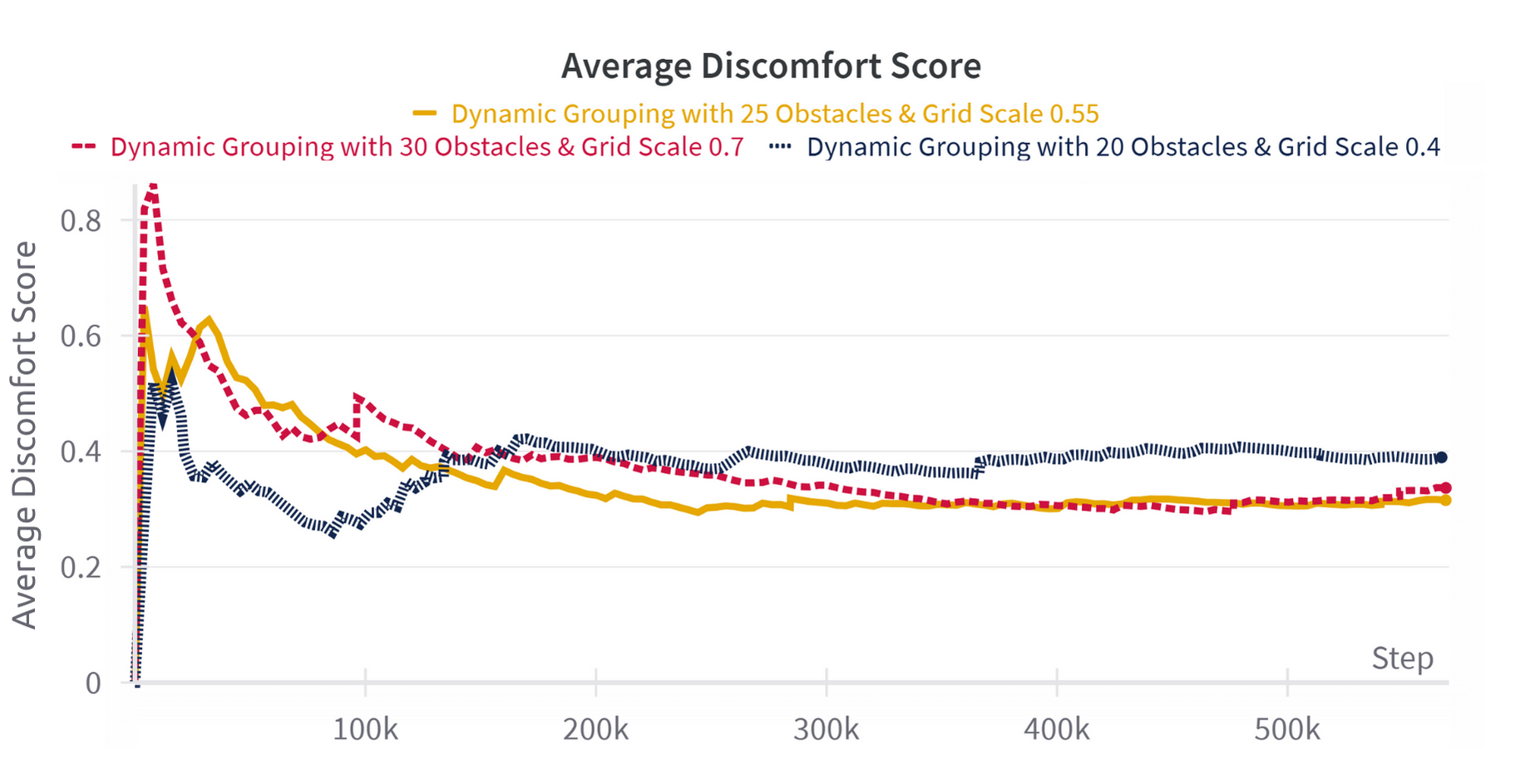}
    \caption{Discomfort Reduction Progression: Discomfort Scores achieved by SANGO agents over 580K training steps in different environments.}
    \vspace{-1.5em}
    \label{fig:SANGO_Discomfort}
\end{figure}
    
\begin{figure}[!h]
    \centering
    \includegraphics[trim={0, 0.425cm, 0, 0.1cm}, clip, width=\linewidth]{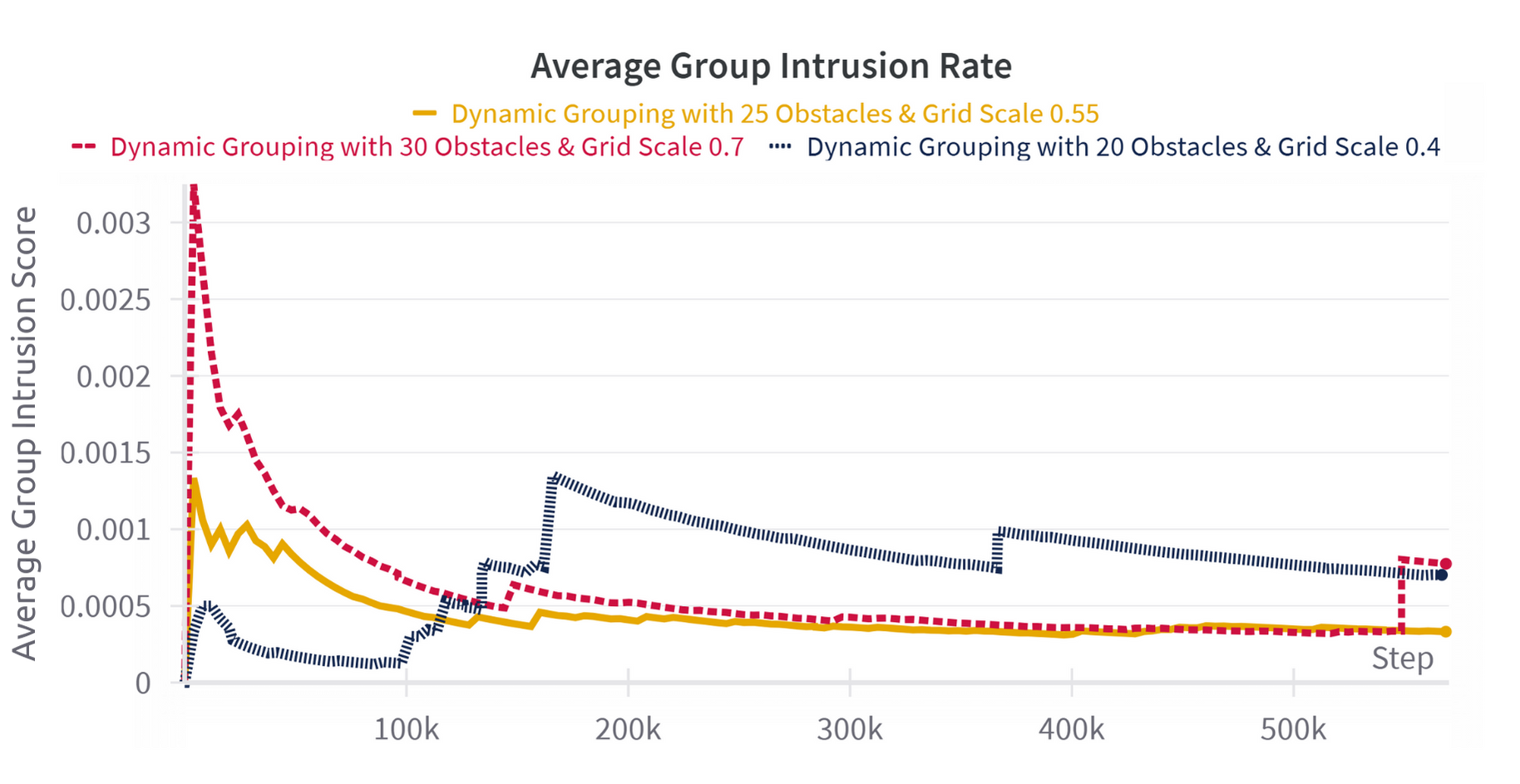}
    \caption{Group intrusion Progression: Group intrusion Rate achieved by SANGO agents over 580K training steps in diverse environments}
    \vspace{-1.1em}
    \label{fig:SANGO_Group_intrusion}
\end{figure}

\begin{table*}[!h]
    \centering
    \caption{Performance of Social Navigation agent, without Dynamic Grouping and with Dynamic Grouping (SANGO), in various complex environments. (All values averaged over 100 test trials)}
    \label{tab:Evaluation}
    \begin{tabular}{|l|c|c|c|c|c|c|}
        \toprule
        \multicolumn{1}{|c|}{Metric (units)} & \multicolumn{2}{|c|}{Environment 1} & \multicolumn{2}{|c|}{Environment 2} & \multicolumn{2}{|c|}{Environment 3}\\
        \hline\
         & Without & With (SANGO) & Without & With (SANGO) & Without & With (SANGO)\\
        \midrule
        Average Discomfort Score & $0.895$ & $0.292$ & $1.779$ & $0.3129$ & $1.994$ & $0.3282$ \\
        Average Group intrusion Rate (\%) & $-$ & $0.04822$ & $-$ & $0.03387$ & $-$ & $0.0535$ \\
        Average Minimum time to Collision (s) & $17.93238$ & $21.02462$ & $16.01446$ & $19.73831$ & $13.78113$ & $17.61131$ \\
        Average Collision Rate (\%) & $5.399$ & $4.995$ & $7.233$ & $5.889$ & $10.242$ & $7.233$ \\
        Average Dynamic Obstacle Collision Rate (\%) & $0.61323$ & $0.0337$ & $0.22845$ & $0.05579$ & $0.29772$ & $0.04515$ \\
        Average Wall \& Obstacle Collision Rate (\%) & $5.383$ & $4.961$ & $7.862$ & $5.933$ & $9.944$ & $7.188$ \\
        Average Timeout & $0.2$ & $0.19$ & $0.35$ & $0.28$ & $0.43$ & $0.23$ \\
        Average Stalled Time (s) & $0.3621$ & $0.3858$ & $0.4382$ & $0.2250$ & $0.5479$ & $0.3177$ \\
        Average Human Distance (m) & $12.654$ & $11.833$ & $9.076$ & $13.637$ & $7.871$ & $13.707$ \\
        Success Rate (\%) & $80$ & $81$ & $65$ & $72$ & $57$ & $77$ \\
        \bottomrule
    \end{tabular}
\end{table*}

\begin{figure}[!h]
    \centering
    \includegraphics[trim={0, 0.425cm, 0, 0.525cm}, clip, width=\linewidth]{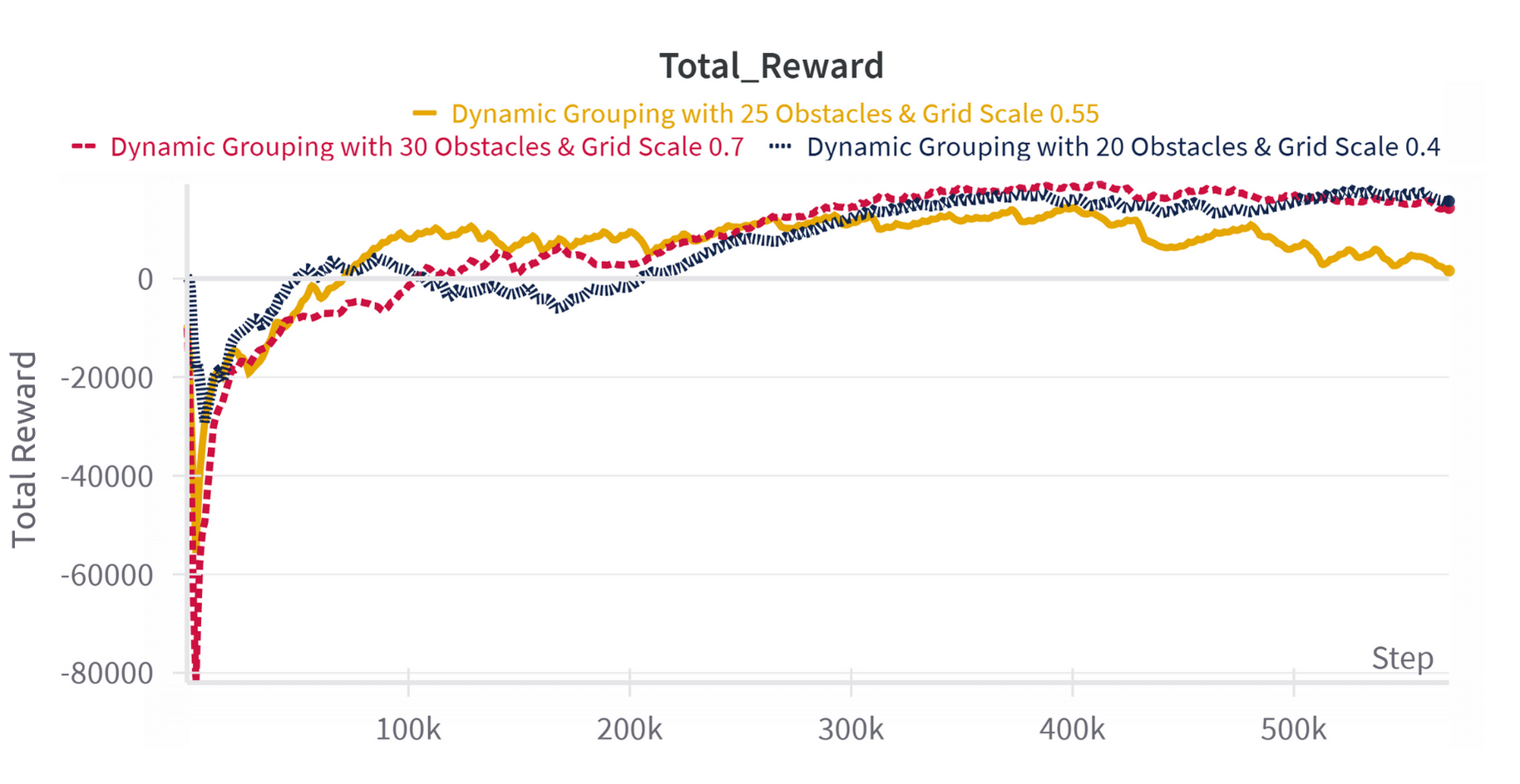}
    \caption{Learning Progression: Total Rewards acquired by SANGO agents during 580K training steps in 3 diverse environments.}
    \vspace{-1.5em}
    \label{fig:SANGO_Rewards}
\end{figure}

\begin{figure}[!h]
    \centering
    \includegraphics[trim={0, 0.425cm, 0, 0.525cm}, clip, width=\linewidth]{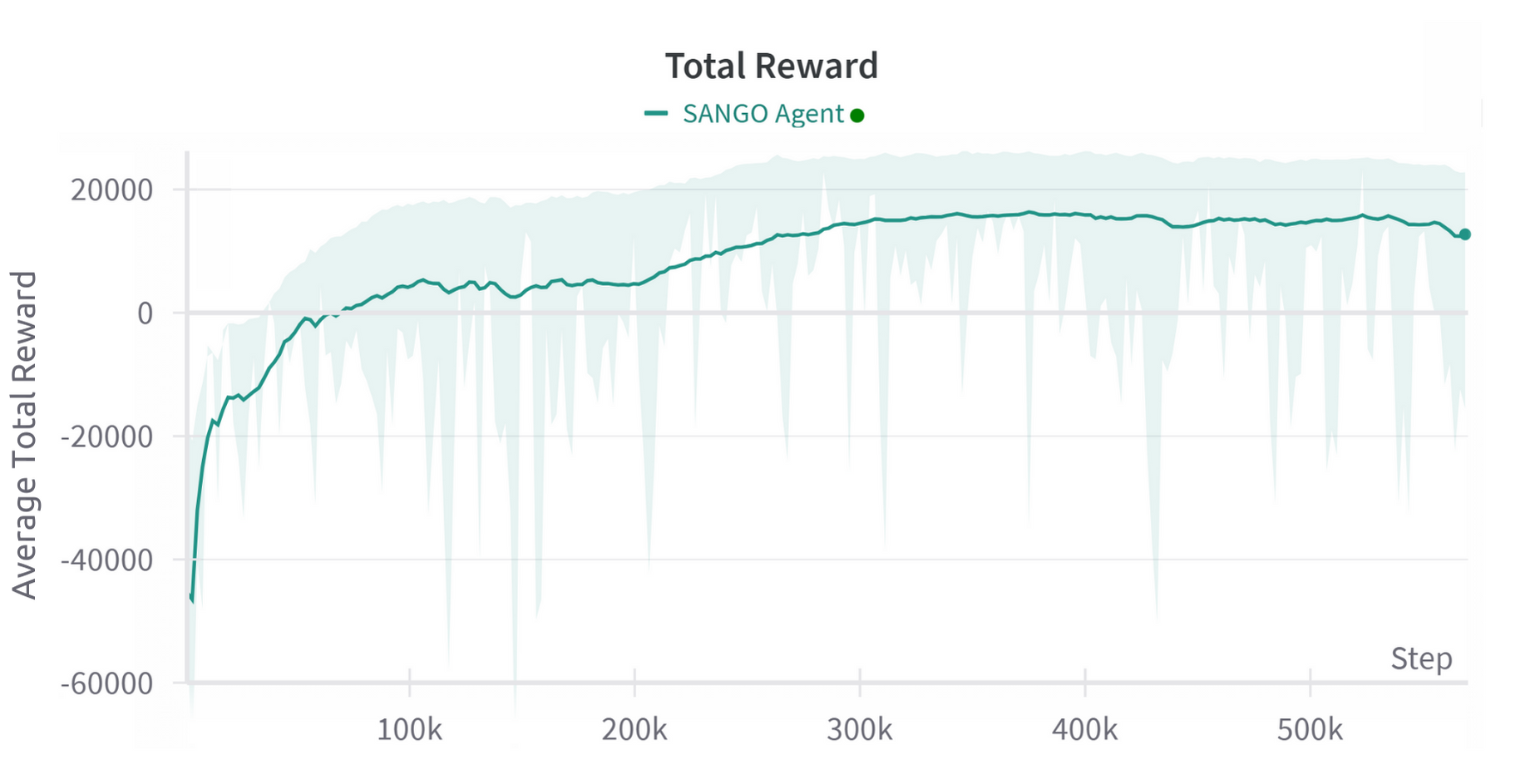}
    \caption{Progression of Learning - Averaged Total Reward obtained by the SANGO agent while training over 580K steps in various environments}
    \vspace{-1.5em}
    \label{fig:SANGO_Total_Rewards}
\end{figure}

\begin{itemize}
    \item \textbf{Average Discomfort Score}: This metric is calculated by averaging the cumulative weighted threshold intrusion and the number of times the agent has crossed the threshold, causing discomfort to groups and dynamic obstacles. The score is a non-negative value, where 0 indicates no discomfort and higher values indicate increasing levels of discomfort. Figure-\ref{fig:SANGO_Discomfort} illustrates the agent's progress in socially aware navigation, showing a reduction in the discomfort score as the number of training steps increases.
    \item \textbf{Average Group intrusion Rate}: The frequency with which the agent intrudes the space of the formed groups. Figure-\ref{fig:SANGO_Group_intrusion} demonstrates the agent's advancement in reducing the average group intrusion rate, highlighting its effectiveness in mitigating group interference.
    \item \textbf{Average Minimum Time to Collision}: The least number of steps taken before a collision occurs (with static and dynamic obstacles, and boundary walls).
    \item \textbf{Average Collision Rate}: The frequency of any collision (with static obstacles, dynamic obstacles, and boundary walls).
    \item \textbf{Average Dynamic Obstacle Collision Rate}: The frequency of collisions with dynamic obstacles.
    \item \textbf{Average Wall and Obstacle Collision Rate}: The frequency of collisions with static obstacles and boundary walls.
    \item \textbf{Average Path Length}: The number of steps taken before reaching the goal position.
    \item \textbf{Total Reward Obtained}: The cumulative reward obtained over the training period, serves as an indicator of the agent's learning progress and a benchmark for the training process. Figure~\ref{fig:SANGO_Rewards} illustrates the agent's learning curve. We also ran experiments on 18 different environments in total with different numbers of dynamic obstacles and in various gyms (MOSANG \& COG). The average aggregated reward obtained in total is illustrated in Figure~\ref{fig:SANGO_Total_Rewards}.
\end{itemize}

All the aforementioned metrics are logged and observed during the training phase, including both episodic metrics and averaged metrics over episodes, to provide a robust understanding of the agent's training and learning capabilities. During the testing phase, these metrics are averaged over multiple episodes.

\subsection{Compute Setup Used}

The training and testing processes were conducted on a single Nvidia RTX A5000 GPU. On average, the training duration spanned approximately 96+ hours with rendered visuals and over 72 hours without renderings, with multiple experiments in various environments running in parallel. This setup provided a robust computational environment to support the intensive training requirements of the PPO algorithm.

\section{RESULTS}

The model, trained for 1 million steps, with optimal performance at 580K steps, was rigorously evaluated across three unseen environments in MOSANG, each with distinct configurations. Environment 1 included 20 dynamic obstacles with a grid scaling of 0.4, Environment 2 featured 25 dynamic obstacles with a grid scaling of 0.55, and Environment 3 had 30 dynamic obstacles with a grid scaling of 0.7. These environments, increasing in complexity, were randomly selected to ensure a comprehensive assessment. This testing phase involved spawning dynamic obstacles and worlds as per the previously detailed methodology. Table~\ref{tab:Evaluation} encapsulates the metrics and performance scores obtained during a test phase spanning 100 episodes. Comparative analyses were conducted with (a) a Socially Aware Navigating agent trained without a grouping mechanism and (b) Our SANGO agent, both in the aforementioned 

The results underscore the superior performance of the agent trained with SANGO, as evidenced by the significant improvements in multiple metrics. The \textit{Average Discomfort Score} for the SANGO agent remains consistently lower compared to the ungrouped agent, indicating a higher level of comfort for dynamic obstacles and groups. Specifically, the scores are 0.292, 0.3129, and 0.3282 for Environment\_1, Environment\_2, and Environment\_3 respectively, compared to 0.895, 1.779, and 1.994 for the ungrouped agent. The \textit{Average Group intrusion Rate} for the SANGO agent remains consistently low, reflecting the agent's ability to navigate without disrupting the integrity of the obstacle groups. In contrast, the ungrouped agent shows a complete lack of intrusion control, as there is no concept of grouping.

In terms of \textit{Average Minimum Time to Collision}, the SANGO agent demonstrates a substantial delay in collision occurrences across all environments, with times of 2102.462, 1973.831, and 1761.131 steps respectively, in contrast to 1793.238, 1601.446, and 1378.113 steps for the ungrouped agent. This indicates a marked improvement in collision avoidance behaviour. The \textit{Average Collision Rate}, \textit{Average Dynamic Obstacle Collision Rate}, and \textit{Average Wall \& Obstacle Collision Rate} are all notably lower for the SANGO agent, highlighting its improved navigation capabilities and reduced collision incidences. Further, the \textit{Average Timeout} and \textit{Average Stalled Time} reveal the efficiency of the SANGO agent in maintaining movement and avoiding unnecessary halts, compared to the ungrouped agent, which exhibits higher timeout and stalled times, particularly in more complex environments.

The \textit{Average Human Distance} metric indicates that the SANGO agent maintains a more consistent and appropriate distance from dynamic obstacles, enhancing its social navigation performance. Finally, the \textit{Success Rate} metric, which reflects the agent's ability to reach its goal, is significantly higher for the SANGO agent across all environments, achieving success rates of 0.81, 0.72, and 0.77 compared to 0.80, 0.65, and 0.57 for the ungrouped agent.

\section{CONCLUSION}

This study introduces SANGO (Socially Aware Navigation through Grouped Obstacles), a novel approach that leverages dynamic obstacle grouping and deep reinforcement learning to adapt the navigation of an agent in a dynamic environment. Our experiments across multiple complex environments demonstrate SANGO's superior performance compared to traditional navigation methods.

Key findings include:
\begin{enumerate}
    \item Significantly reduced discomfort scores (by up to 83.5\% in the most complex environment), indicating SANGO's ability to navigate while respecting social norms and personal space.
    \item Consistently lower collision rates across all obstacle types, with improvements of up to 29.4\% in overall collision avoidance.
    \item Enhanced navigation efficiency, as evidenced by increased minimum time to collision (up to 27.8\% improvement) and reduced stalled time (up to 48.7\% reduction in complex scenarios).
    \item Improved success rates in reaching the goal, with up to 35\% higher success rates in the most challenging environments.
    \item Maintenance of appropriate distances from dynamic obstacles, demonstrating SANGO's capacity for socially aware positioning.
\end{enumerate}

SANGO's performance improvements are particularly pronounced in complex, densely populated environments, suggesting its potential for real-world applications in crowded spaces such as hospitals, airports, or shopping centres. The DBSCAN algorithm's role in dynamic obstacle grouping proves crucial, enabling the agent to navigate around cohesive groups rather than individual obstacles.

Although the SANGO model demonstrates significant advancements in socially aware navigation, several challenges remain. Currently, dynamic obstacle motion is modelled in 2D using known social motion algorithms. Expanding this to 3D environments and incorporating real-world human trajectory data along with multi-modal inputs could improve generalization from simulation to real-world tasks. Additionally, the model's reliance on positional cues limits social compliance, which could be improved by integrating more complex social norms into the reward function. Finally, the model works well in predefined environments, but adaptive learning mechanisms are necessary to handle novel or rapidly changing scenarios. Addressing these limitations will enhance the scalability of the model and its effectiveness in the real world.

In conclusion, SANGO marks a substantial improvement in socially aware navigation by integrating dynamic obstacle handling with deep reinforcement learning, resulting in better navigation performance and social awareness.

\bibliographystyle{IEEEtran}
\bibliography{references}

\end{document}